\documentclass[runningheads]{llncs}
\usepackage[T1]{fontenc}
\usepackage{tcolorbox}
\usepackage{hyperref}
\usepackage{graphicx} % KB
\usepackage{longtable} % KB
\definecolor{backcolour}{RGB}{225,225,247}
\definecolor{keywordsblue}{RGB}{0,0,204}
\definecolor{string}{RGB}{72,94,70}
\definecolor{commentgreen}{RGB}{2,112,10}
\definecolor{symbolpurple}{RGB}{153,0,153}
\definecolor{codegray}{RGB}{96,96,96}
\usepackage[font=small,labelfont=bf]{caption}
\usepackage{listings}

\lstdefinelanguage{ASTRA}{
    keywords=[1]{package, agent, extends, rule, goal, query, true, false, 
        explain, formula, module, import, try, recover, inference, initial, 
        send, foreach, forall, \(, \), learn, algorithm, types, message, if,
        else, constant, wait, while},
    sensitive=false, % keywords are not case-sensitive
    keywordstyle=[1]\color{keywordsblue},  
    commentstyle=\color{commentgreen},
    numberstyle=\tiny\color{codegray},
    stringstyle=\color{string}\ttfamily,
    basicstyle=\footnotesize,
    breakatwhitespace=false,         
    breaklines=true,                 
    captionpos=t,
    frame=b,
    belowcaptionskip=10pt,
    framextopmargin=10pt,
    keepspaces=true,                 
    numbers=left,                    
    numbersep=5pt,                  
    showspaces=false,                
    showstringspaces=false,
    showtabs=false,                  
    tabsize=2,
    stepnumber=1,
    numberfirstline=false,
    firstnumber=1,
    string=[b]{'},
    morecomment=[l]{//}, % l is for line comment
    morestring=[b]", % defines that strings are enclosed in double quotes
    morekeywords=[2]{>,<,\{,\},-,!,=,~,+,@message,\#,\$,list,int,string,double,boolean,::},
    otherkeywords={>,<,\{,\},-,!,=,~,+,@message,\#,\$},
    keywordstyle=[2]\color{symbolpurple},
    numberstyle=\tiny, 
    basicstyle=\scriptsize
} 

\DeclareCaptionFormat{listing}{% 
  \hspace*{-0.4pt}\colorbox{gray!30}{\parbox{\dimexpr\textwidth-2\fboxsep+0.9pt\relax}{#1#2#3}}} 
\NewDocumentCommand \mod {m} {\textsf{\texttt{#1}}}
\begin{document}
\title{\mod{astra-langchain4j}: Experiences Combining LLMs and Agent Programming}
\author{
Rem Collier\orcidID{0000-0003-0319-0797}\inst{1}\and
Katharine Beaumont\orcidID{0009-0001-9250-0090}\inst{1}\and
Andrei Ciortea\orcidID{0000-0003-0721-4135}\inst{2}}

\authorrunning{R. Collier et al.}
\institute{School of Computer Science, University College Dublin, Ireland
\email{rem.collier@ucd.ie, katharine.beaumont@ucdconnect.ie}\and
School of Computer Science, University of St.Gallen, St. Gallen, Switzerland
\email{andrei.ciortea@unisg.ch}\\
}
\maketitle              % typeset the header of the contribution
\begin{abstract}
Given the emergence of Generative AI over the last two years and the increasing focus on Agentic AI as a form of Multi-Agent System it is important to explore both how such technologies can impact the use of traditional Agent Toolkits and how the wealth of experience encapsulated in those toolkits can influence the design of the new agentic platforms. This paper presents an overview of our experience developing a prototype large language model (LLM) integration for the ASTRA programming language. It presents a brief overview of the toolkit, followed by three example implementations, concluding with a discussion of the experiences garnered through the examples.
\end{abstract}
\section{Introduction}
\label{sec:intro}

% Need to introduce references and related work briefly here...
The last couple of years have seen a spike in interest in the concept of agents that has not been seen since the late 1990s. The source of the interest is centred around the use of Large Language Models (LLMs) \cite{patil2024review} and Generative Artificial Intelligence (GenAI) \cite{gozalo2023chatgpt} to implement reasoning and planning for goal- and task-oriented agents, in an area that has become known as Agentic AI \cite{acharya2025agentic}.

To date, there has been only limited work exploring how mainstream agent toolkits can take advantage of Generative AI and how Generative AI can benefit from those toolkits \cite{thisVolumeBriolaEtAl}. This paper presents details of our recent experiences in integrating support for LLMs into the ASTRA\cite{collier2015reflecting} agent programming language. Our integration takes the form of a library built on top of LangChain4J\footnote{\url{https://github.com/langchain4j/langchain4j}}, an open-source Java library for integrating LLMs. The library is known as \mod{astra-langchain4j}, has been part of ASTRA\footnote{\url{https://gitlab.com/astra-language/astra-core}} since version 2.0.3, and is freely available from Maven Central\footnote{\url{https://central.sonatype.com/artifact/com.astralanguage/astra-langchain4j}}. To support the use of LLMs, we introduce mechanisms for creating prompts and for processing responses. We also introduce the idea of \textit{BeliefRAG} --- a simple mechanism that allows the agents' beliefs to be mined for information that can be used to supplement a prompt.

The \mod{astra-langchain4j} library is introduced in Section \ref{sec:library}. It is then followed in Section \ref{sec:examples} by an overview of some of the sample systems we have developed with our library which we will demonstrate if the paper is accepted.  Finally, Section~\ref{sec:experiences} summarises some of our early experiences in using LLMs with ASTRA.

\begin{figure*}[!b]
    \centering
    \includegraphics[width=\linewidth]{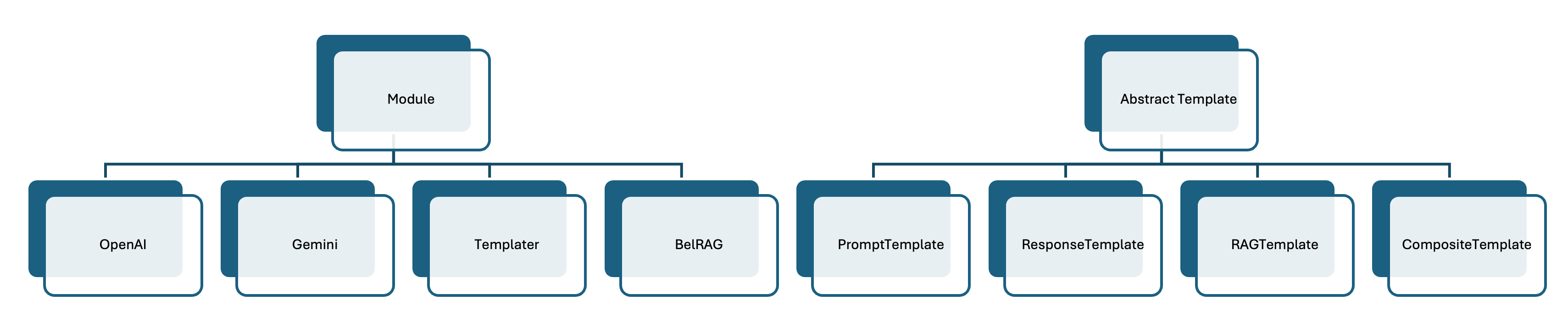}
    \caption{The modules and templates provided by \mod{astra-langchain4j}}
    \label{fig:hierarchies}
\end{figure*}

\section{The \mod{astra-langchain4j} library}
\label{sec:library}
ASTRA incorporates the concepts of a module as a mechanism to allow developers to add their own custom sensors, actions (actuators), predicates, terms and events through the concept of a module \cite{collier2015reflecting}. The first two of these features are well established in the literature: sensors allow the polling of the environment to monitor for changes and the updating of the agents' beliefs to reflect those changes, and actions provide the primitive capabilities of the agent. The next two features are less traditional, but are also widely used in languages like Jason~\cite{bordini2007programming}: predicates allow for custom formulae that can be used in the context of an agents plans (for example, checking if a string contains a substring) and terms can be used to generate values that can be used in predicates (for example, calculating the length of a string). The final feature is less common in AgentSpeak(L) style languages: it allows the creation of custom events that can be added to the agents event queue and used to directly trigger plans. In ASTRA, modules are used extensively to support, for example, integrations with environment frameworks like the Environment Interface Standard~\cite{behrens2011towards} and CArtAgO~\cite{ricci2009environment}, support for incorporating Swing UIs and external tools like Message Brokers, and the provision of basic functions for manipulating strings, lists, console output. In ASTRA, a library is simply a collection of modules developed for a specific purpose.

This section introduces \mod{astra-langchain4j}, a new ASTRA library built on Langchain4J, a Java library to implement LLM-based systems.
The library brings the following capabilities to agent programs written in ASTRA: (i) executing LLM calls, (ii) defining, using, and reusing prompt templates, and (iii) exploiting an agent's belief base for retrieval-augmented generation (RAG).
The source code is available as part of the ASTRA codebase\footnote{\url{https://gitlab.com/astra-language/astra-core/-/tree/master/astra-langchain4j}}. Compiling and running ASTRA code is integrated into the Maven build system\footnote{\url{https://maven.apache.org/}} and the library can be used in any ASTRA project by adding the relevant Maven dependency.

A summary of the main classes provided as part of the library can be seen in Figure \ref{fig:hierarchies}. The hierarchy on the left-hand side identifies four modules that have been provided. This includes two modules that provide sample integrations with LLMs and two modules that provide basic support for creating templates that can be used to interact with the LLMs.

The LLM integrations provided are for OpenAI's ChatGPT\footnote{\url{https://chatgpt.com/}} and Google Gemini\footnote{\url{https://gemini.google.com/}} and are called \verb|OpenAI| and \verb|Gemini| respectively. These integrations are very simple and are adapted from the getting started documentation provided with Langchain4j\footnote{\url{https://docs.langchain4j.dev/get-started}}. Each module contains a number of actions that can be used to interact with the LLM. These actions are standardised across the modules for consistency. To make their use easier, two ASTRA agent classes are provided to simplify their incorporation into larger programs. For example, Listing \ref{code:openai} provides the basic integration for OpenAI-based LLMs.

\begin{lstlisting}[float=!t,caption={OpenAI Base Class},language=ASTRA,label=code:openai]
package astra.langchain4j;

import astra.langchain4j.OpenAI;

agent OpenAIAgent {
    module OpenAI model;

    rule +!initialize(string key) {
        model.loadApiKeyFromFile(key);
        model.initialize();
    }

    rule +!initialize(string key, list params) {
        model.loadApiKeyFromFile(key);
        model.initialize(params);
    }
}
\end{lstlisting}

\begin{lstlisting}[caption={Example of minimal agent program},language=ASTRA,label=code:integration]
import astra.langchain4j.OpenAIAgent;

agent MyAgent extends OpenAIAgent {
    rule +!main(list args) {
        OpenAIAgent::!initialize("../api.key").
    }
}
\end{lstlisting}

Here, there are two initialization plans: the first has a \verb|key| parameter, which is a relative path to a file containing an OpenAI key, and the second adds another parameter which is a list of \verb|params| that can be used to pass configuration parameters when connecting to the OpenAI service (e.g. temperature, topK, ...). As is shown in Listing \ref{code:integration}, this can be easily incorporated into any program using the multiple inheritance reuse mechanism provided with ASTRA \cite{dhaon2014multiple}.

\begin{lstlisting}[float=t!,caption={A Joker Agent},language=ASTRA,label=code:joke]
import astra.langchain4j.*;

agent Joker extends OpenAIAgent {
    module Console C;
    module Templater templater;
    
    rule +!main(list args) {
        OpenAIAgent::!initialize("../api.key").

        PromptTemplate template = templater.createPromptTemplate("
            why did the ${animal} cross the road?
        ");
        tpl.addBinding(template, "animal", "hedgehog");
        model.chat(template, string reply);
        C.println("Response: " + reply);
    }
}
\end{lstlisting}

\begin{lstlisting}[float=b!,caption={A Happy Agent},language=ASTRA,label=code:answer]
import astra.langchain4j.*;

agent Happy extends OpenAIAgent {
    module Console C;
    module Templater templater;
    
    rule +!main(list args) {
        OpenAIAgent::!initialize("../api.key").

        string input = "
            Are you happy? Answer YES or NO only using
            the template '**Result <answer>**'
        ";
        model.chat(input, string reply);
        ResponseTemplate template =
            templater.createResponseTemplate("Result **${answer}**");
        templater.inferBindings(template, reply);
        string answer = templater.getBinding(response, "answer");

        C.println("Response: " + answer);
    }
}
\end{lstlisting}

The templating modules provide support for generating prompts and parsing responses. The set of templates provided in our initial solution are highlighted in the class hierarchy on the right-hand side of Figure \ref{fig:hierarchies}. The templates provide a way of formulating an input to an LLM that can be parameterised as is appropriate. In the text, parameters are encapsulated in a special token: \verb|${<param>}| for example, \verb|Why did the ${animal} cross the road?| captures the classical joke, providing a parameter \verb|animal| that can be used to change the focus of the joke. The common code for all of the templates is implemented in the \verb|AbstractTemplate| class. The \verb|PromptTemplate| class provides basic prompts like the joke provided above (see Listing \ref{code:joke}). 

The templates are designed to be reusable, and can be \verb|reset()| via the \verb|Templater| module, removing any variable bindings associated with the template. To process responses, a \verb|ResponseTemplate| is also provided. This also uses parameters but also incorporates the ability to infer parameter bindings from the response provided. The example in Listing \ref{code:answer} illustrates this.

The \verb|RAGTemplate| is created through the \verb|BelRAG| module and provides a basic form of Retrieval Augmented Generation (RAG)\cite{lewis2020retrieval} that allows the agent to incorporate knowledge stored in its beliefs into a chat prompt. Listing \ref{code:foodie} below demonstrates how the approach works with a simple example where the LLM is asked to identify which items on a list are fruits.

\begin{lstlisting}[float=b!,caption={A Food Agent},language=ASTRA,label=code:foodie]
import astra.langchain4j.*;

agent Foodie extends OpenAIAgent {
    module Console C;
    module Templater templater;
    module BelRAG rag;
    
    types eg {
        formula food(string);
    }

    initial food("nuts"), food("apples"), food("oranges");
    
    rule +!main(list args) {
        OpenAIAgent::!initialize("../api.key").
        
        RAGTemplate ragTemplate = rag.createRAGTemplate("
            Which of the following are fruits?
        ");
        rag.addInput(ragTemplate, food(string A), "${A}");
        
        model.chat(ragTemplate, string reply);

        C.println("Reply: " + reply);
    }
}
\end{lstlisting}

In the example, the RAGTemplate is created on lines 17-20 and it combines a piece of introductory text with a set of inputs (here there is only one on line 20). The inputs map predicates (in this case \verb|food(string A)|) to text where the parameters in the predicate are also parameters in the text (in this example, the text is simply the food). The input is generated at the point that the chat message is sent to the LLM (here line 22).  This involves the module querying the current beliefs of the agent and generating a set of input statements that are combined with the introductory text and then sent to the LLM.

The final template mechanism provided is a composite mechanism that allows the developer to combine basic prompts with RAG prompts as is appropriate. We see an example of this in Section \ref{sec:towerworld}.

\section{Example Systems}
\label{sec:examples}

At this point, three example systems have been created using the \mod{astra-langchain4j} library. All can be downloaded from our Gitlab repository\footnote{\url{https://gitlab.com/astra-language/examples/llm-examples}}. Each of these examples is explored in more detail in the sections below.

\begin{lstlisting}[float=b!,caption={The Main agent},language=ASTRA,label=code:planner]
agent Main extends RoundRobin {
    initial
        role("planner",
            "A helpful assistant that can plan trips.",
            "You are a helpful assistant that can suggest a travel
            plan for a user based on their request."),
        role("local",
            "A local assistant that can suggest local activities
             or places to visit.",
            "You are a helpful assistant that can suggest authentic 
            and interesting local activities or places to visit for 
            a user and can utilize any context information provided."),
        role("language",
            "A helpful assistant that can provide language tips for
            a given destination.",
            "You are a helpful assistant that can review travel 
            plans, providing feedback on important/critical tips about
            how best to address language or communication challenges 
            for the given destination. If the plan already includes
            language tips, you can mention that the plan is
            satisfactory, with rationale."),
        role("summary",
            "A helpful assistant that can summarize the travel plan.",
            "You are a helpful assistant that can take in all of the
            suggestions and advice from the other agents and provide
            a detailed final travel plan. You must ensure that the 
            final plan is integrated and complete. YOUR FINAL RESPONSE
            MUST BE THE COMPLETE PLAN. When the plan is complete and 
            all perspectives are integrated, you can respond with 
            TERMINATE."),
        order(["planner", "local", "language", "summary"]);

    rule +!main(list args) {
        !setup();
        !plan("Plan a 3 day trip to Nepal.");
    }
}
\end{lstlisting}

\begin{lstlisting}[float=t!,caption={Sketch of the Assistant agent},language=ASTRA,label=code:assistant]
agent Assistant extends Common {
    rule +!main([string description, string systemMessage]) {
        +description(description);
        +systemMessage(systemMessage);
        oai.loadApiKeyFromFile("../api.key");
        oai.initialize("gpt-4o");
    }

    rule @message(request, string sender, task(string task)) :
            description(string desc) & systemMessage(string sm) {
        send(agree, sender, task(task));
        
        PromptTemplate template = tpl.createPromptTemplate(
            "${description}${systemMessage}${task}"
        );
        tpl.addBinding(template, "description", desc);
        tpl.addBinding(template, "systemMessage", sm);
        tpl.addBinding(template, "task", task);

        oai.chat(template, string reply);
        send(inform, sender, result(reply));
    }
}
\end{lstlisting}

\subsection{Travel Planner}
\label{sec:travel}

The first example developed is a \textit{Travel Planner} example which is based on an example provided as part of the AutoGen documentation\footnote{\url{https://microsoft.github.io/autogen/stable/user-guide/agentchat-user-guide/examples/travel-planning.html}} that consists of a team of 4 specialised agents that collaborate with one another to provide a comprehensive travel itinerary.  Each of the specialists are instances of an \verb|AssistantAgent| class and collaboration takes place via a \verb|RoundRobinGroupChat|. To replicate this example with \mod{astra-langchain4j}, we choose to use the FIPA Request Interaction Protocol\footnote{\url{http://www.fipa.org/specs/fipa00026/}} for orchestrator-assistant interaction and create two simple agents: an \verb|Assistant| agent and the \verb|Main| agent. The latter acts as the orchestrator, and implements the Travel Planning scenario via the code presented in Listing~\ref{code:planner}: each of the four roles in this scenario (i.e., \texttt{planner}, \texttt{local}, \texttt{language}, and \texttt{summary}) is associated with a description and a system message corresponding to the information provided in the AutoGen example.

The code for the \verb|Assistant| agent is given in Listing \ref{code:assistant}. In the code, the \verb|!main()| plan is called when the agent is created, with two strings being passed: a description (\verb|desc|) and a system message (\verb|sm|). %, which correspond to the information provided in the AutoGen example.
The second plan relates to the receipt of an incoming request from another agent (the orchestrator) asking the agent to perform a given task.  The agent responds as per the protocol by first agreeing to perform the task, then executing the task (lines 9-22) and finally informing the sender of the result of the task.

\begin{lstlisting}[float=t!,caption={Sketch of the Orchestrator (RoundRobin) agent},language=ASTRA,label=code:round]
agent RoundRobin extends Common {
    rule +!setup() {
        foreach(role(string name, string desc, string sm)) {
            S.createAgent(name, "Assistant");
            S.setMainGoal(name, [desc, sm]);
        }
    }
    rule +!plan(string task) : order(list names) {
        string mess = task;
        forall(string name : names) {
            send(request, name, task(mess));
            wait(responded(name, string result));
            C.println("-----" + name + "-----");
            C.println(result);
            mess = mess + task;
        }
    }
    rule @message(inform, string sender, result(string result)) {
        +responded(sender, result);
    }
}
\end{lstlisting}

The orchestrator (\verb|Main|) agent code is outlined in the \verb|RoundRobin| agent in Listing \ref{code:round}. This agent contains three plans. The first creates a set of \verb|Assistant| agents based on the available \verb|role(...)| beliefs, where each agent is initialised with a description and system message. The second plan implements the travel planning task, which triggers the FIPA Request protocol by sending a \verb|request| message to each of the role-playing agents in a specified order (based on the \verb|order(...)| belief).  The requests are sent sequentially, with the agent waiting for each role to respond (the \verb|wait| statement) before starting the next step. While the initial task request contains only the initial task string, at the end of the step, this string is combined with the response of the first agent and then sent to the next agent. This is repeated for each agent, ensuring that the next agent has full knowledge of what was said in the previous step. This allows the agents to use the previous messages in the conversation to construct their response. The final plan handles the response from each \verb|Assistant|, adopting a belief that triggers the next interaction.
% Finally, the \verb|Main| agent implements the Travel Planning scenario via the code presented in Listing \ref{code:planner}.

\subsection{Tic-Tac-Toe}
\label{sec:ttt}

The second example developed is an implementation of Tic-Tac-Toe where two players interact with a simple web-based Tic-Tac-Toe server.  This example was originally developed to illustrate the Multi-Agent MicroServices (MAMS) framework \cite{o2020delivering,w2019mams}. The agent that plays the game is implemented in a common base agent program called \verb|AbstractPlayer|. The full code for this agent program is not presented here, but can be found in Gitlab\footnote{\url{https://gitlab.com/astra-language/examples/llm-examples/tic-tac-toe/-/blob/main/2llms/src/main/astra/AbstractPlayer.astra}}. In a nutshell, it implements the interactions between the player and the Tic-Tac-Toe server, with the state of the board being stored in a \verb|board(...)| belief that contains a single argument that is an instance of the \verb|JsonNode| Java class that is provided as part of the popular FasterXML/Jackson JSON library\footnote{\url{https://github.com/FasterXML/jackson}}. The game play also includes a simple protocol where each player takes their turn and then sends a FIPA ACL \verb|request| to the other player to make their move. The decision around what move to make is encoded as a test goal \verb|?location(int row, int column)|. To evaluate the use of LLMs in playing Tic-Tac-Toe, a number of player agent types have been created that implement a plan to achieve this test goal. We start in Listing \ref{code:linear} with the \verb|BasicPlayer| agent, a non-LLM player that is used as a baseline for evaluating to LLM-based approaches. This agent is a \textit{linear player}, in that it places its token in the first available space on the board. It uses a \verb|location(...)| belief to investigate which cells contain a value.

\begin{lstlisting}[float=t!,caption={The BasicPlayer agent},language=ASTRA,label=code:linear]
agent BasicPlayer extends AbstractPlayer {
    rule +?location(int row, int column) {
        row = -1;
        int i=0;
        while (i<3 & row == -1) {
            int j=0;
            while (j<3 & row == -1) {
                if (location(i, j, "")) {
                    row = i;
                    column = j;
                }
                j++;
            }
            i++;
        }
    }
}
\end{lstlisting}

The first attempt to create a basic LLM player involved asking an LLM to select the cell as is shown in Listing \ref{code:oai-basic}. The code presented is a sketch of the solution, which proved to be bad at playing, and was consistently beaten by the Basic non-LLM player. It had no awareness of being in a loosing position and did not seem to adopt a consistent or coherent winning strategy. At times, the LLM would recommend a location that had already been played. We accept that this may be a result of the prompt used, but understand that this is consistent with other researchers findings related to LLMs playing board games \cite{topsakal2024evaluating}.

\begin{lstlisting}[float=t!,caption={Outline of the Basic OpenAI Player agent},language=ASTRA,label=code:oai-basic]
agent OpenAIPlayer extends OpenAIAgent, AbstractPlayer {
    rule +?location(int row, int column) :
            token(string token) & board(JsonNode b) {
        PromptTemplate template = templater.createPromptTemplate("
            if the following json is a representation of a tic-tac-toe
            board ${board}, what is the best move player '${player}'
            can make? 
            Answer in the form '**Play ${player} at <X>, <Y>**'
        ");
        templater.addBinding(template,"board",builder.toJsonString(b));
        templater.addBinding(template, "player", token);

        model.chat(template, string reply);

        ResponseTemplate response = templater.createResponseTemplate(
            "**Play ${player} at ${x}, ${y}**"
        );
        templater.addBinding(response, "player", token);
        templater.inferBindings(response, reply);
        row = M.intValue(templater.getBinding(response, "x"));
        column = M.intValue(templater.getBinding(response, "y"));
    }    
}
\end{lstlisting}

As a test, we replaced the OpenAI LLM we were using, \verb|chatgpt-4o-mini|, with Google Gemini\footnote{\url{https://gemini.google.com/}}, specifically, \verb|gemini-1.5-flash|. The resulting agent program is called \verb|GeminiPlayer|, but this performed no better. Based on our initial results, we investigated two more complex strategies. The first strategy explored was to implement a solution based on Anthropic's \textit{Evaluator-Optimizer} workflow \cite{anthropic}, and the second strategy involved asking the agent to first assess the state of the board to see if it is in a winning or loosing position.

The \textit{Evaluator-Optimizer} solution was no better than the basic LLM player. The source code for this agent is the \verb|ReflectiveOpenAIPlayer| program which can be found in Gitlab\footnote{\url{https://gitlab.com/astra-language/examples/llm-examples/tic-tac-toe/-/blob/main/2llms/src/main/astra/ReflectiveOpenAIPlayer.astra}}. Basically, the program involves two prompts: the first is the same as \verb|OpenAIPlayer|, and the second asks the LLM to assess whether or not the proposed move is a good move. If it judges the move to be bad, the initial prompt is re-run with an additional component that lists the previous moves and asks the LLM not to pick the same move again. The response of the second prompt was inconsistent, with the LLM, on occasion, viewing a blocking move that would stop the opponent winning as being not a good move. At times, the agent would get stuck in a loop where it rejected all available moves (even though there was at least one good move).

A sketch of the final attempt can be seen in listing \ref{code:defensive}. This agent first asks the LLM to decide if the board is in a loosing position for the player. Based on the response returned the agent chooses a second prompt to select a move either to not loose the game or to win the game. This solution provided some competition for the linear player and at one point was beating it consistently. However, when the same code was rerun on a following day, it again started loosing to the linear player.

\newpage
\begin{lstlisting}[caption={Sketch of the Defenive OpenAI Player agent},language=ASTRA,label=code:defensive]
agent DefensiveOpenAIPlayer extends OpenAIAgent, AbstractPlayer {
    rule +?location(int row, int column) : 
            token(string token) & board(JsonNode b) {
        PromptTemplate template;
        ?loosable(board, token, string answer);
        if (answer == "YES")
            template = templater.createPromptTemplate("
                I am a tic-tac-toe playing agent. 
                if the following json is a representation of a 
                tic-tac-toe board ${board}, what location should 
                player '${player}' select to not loose the game?
                Answer in the form '**Play ${player} at <X>, <Y>**'
            ");
        else 
            template = templater.createPromptTemplate("
                I am a tic-tac-toe playing agent. 
                if the following json is a representation of a 
                tic-tac-toe board ${board},what location should 
                player '${player}' select? 
                Answer in the form '**Play ${player} at <X>, <Y>**'
            ");
        templater.addBinding(template,"board",builder.toJsonString(b));
        templater.addBinding(template,"player",token);

        model.chat(template, string reply);

        ResponseTemplate response = templater.createResponseTemplate(
            "**Play ${player} at ${x}, ${y}**"
        );
        templater.addBinding(response, "player", token);
        templater.inferBindings(response, reply);
        row = M.intValue(templater.getBinding(response, "x"));
        column = M.intValue(templater.getBinding(response, "y"));
    }
    rule +?loosable(JsonNode b, string token, string answer) {
        PromptTemplate template = templater.createPromptTemplate("
            I am a tic-tac-toe playing agent. if the following json 
            is a representation of a tic-tac-toe board ${board}, 
            and I am player ${player}. Can I lose the game? Answer 
            YES or NO only using the template '**Result <answer>**'
        ");
        templater.addBinding(template,"board",builder.toJsonString(b));
        templater.addBinding(template, "player", token);

        model.chat(template, string reply);

        ResponseTemplate response = templater.createResponseTemplate(
            "**Result ${answer}**"
        );
        templater.inferBindings(response, reply);
        answer = templater.getBinding(response, "answer");
    }
}
\end{lstlisting}

\subsection{Towerworld}
\label{sec:towerworld}

\begin{lstlisting}[float=t!,caption={Sketch of the Prompt for the Tower agent},language=ASTRA,label=code:tower]
agent Main extends OpenAIAgent { 
    rule +!main(list args) {
        OpenAIAgent::!initialize("../api.key",[temperature(0.0)]);

        // Create & Store RAGModel...
        CompositeTemplate compositeTemplate = 
            templater.createCompositeTemplate();

        PromptTemplate promptTemplate = templater.createPromptTemplate("
            I am an agent that can build block towers.  A block tower is 
            formed by placing blocks so they sit on top of each other.
            The bottom block of the tower sits on the table. Towers are 
            defined as a list of blocks where blocks have names like 
            \"a\", \"b\" or \"c\". The first block in the list sits on
            the table. 
            
            In the situation where I want to build the tower
            [\"a\", \"b\"] then the desired state of the table would be:

            block \"a\" is on the table and block \"b\" is on \"a\". 
            Block names should be lowercase.
        ");
        templater.addTemplate(compositeTemplate, promptTemplate);

        RAGTemplate ragTemplate = rag.createRAGTemplate("
            The following sentences define the current state of the
            blocks.
        ");
        rag.addInput(ragTemplate, 
            on(string A, string B), "block ${A} is on top of ${B}.");
        rag.addInput(ragTemplate,
            holding(string C), "the gripper is holding ${C}.");
        templater.addTemplate(compositeTemplate, ragTemplate);

        promptTemplate = templater.createPromptTemplate("
            I can perform two actions: 
            - if I am not holding anything, I can pick up a block 
              using the json \"{ 'action':'pickup', 'args':[X] }\". 
            - if I am holding a block A, then I can put A down on 
              either another block or the table using the json 
              \"{ 'action':'putdown', 'args':[A, B] }\" 
              where B is either the name of a block or the 'table'. 
            
            Given the current state of the blocks, what sequence of 
            actions should be performed next to build the tower 
            ${tower}?

            Answer with only the list of actions defined using JSON.
        ");
        templater.addTemplate(compositeTemplate, promptTemplate);
        +template(compositeTemplate);

        ei.launch("hw","dependency/tower-1.4.0.jar");
        ei.init();
        ei.start();

        ei.link(GRIPPER);
    }
\end{lstlisting}

The final example explored is \textit{Towerworld}, a classical AI problem available through the Environment Interface Standard (EIS)\cite{behrens2011towards}. For this example, the objective was to use an LLM to build towers of blocks. As a baseline, a standard ASTRA implementation was used\footnote{\url{https://gitlab.com/astra-language/examples/styles/towerworld}}. Initial efforts focused on a similar strategy to that used in the Tic-Tac-Toe example (Section \ref{sec:ttt}). Specifically, the agent was given a prompt that presented the current state of the environment, the goal state (a tower configuration) and a description of the possible actions. The LLM was asked to select the next action based on this prompt.  The results were bad --- the agent was sometimes able to work out what block to first pick up, but then struggled to select an appropriate second move.  In the end, the more successful strategy, which was adopted, was to ask the LLM to define a sequence of actions that should be performed to get from the initial state to the goal state.

\begin{lstlisting}[float=t!,caption={Sketch of the implementation for the Tower agent},language=ASTRA,label=code:tower2]
    rule +$ei.event(block("c")) { +target(["a", "b", "c"]); }
    
    rule +target(list L) { !tower(L); S.exit(); }

    rule +!tower(list L) : template(CompositeTemplate template) {
        templater.reset(template);
        templater.addBinding(template, "tower", P.toString(L));
        model.chat(template, string reply);

        ResponseTemplate response =
            templater.createResponseTemplate("```json${json}```");
        templater.inferBindings(response, reply);

        JsonNode node = converter.parse(
            templater.getBinding(response, "json")
        );

        int i = 0;
        while (i < converter.length(node, "/")) {
            !do(converter.getNode(node, "/["+i+"]"));
            i=i+1;
        }
    }

    rule +!do(JsonNode node) :
            converter.compare(node,"action","pickup") {
        string blk=lc.lower(converter.valueAsString(node, "args/[0]"));
        ei.pickup(blk);
        wait(holding(blk));
    }

    rule +!do(JsonNode node) :
            converter.compare(node,"action","putdown") {
        string X = lc.lower(converter.valueAsString(node, "args/[0]"));
        string Y = lc.lower(converter.valueAsString(node, "args/[1]"));
        ei.putdown(X, Y);
        wait(on(X, Y));
    }
}
\end{lstlisting}

A sketch of the code used to create the prompt for this solution can be found in Listing \ref{code:tower}. This example uses a \textit{composite prompt} that combines basic prompts that describe how the environment is represented with a \verb|RAGTemplate| prompt that is used to extract information about the state of the environment from the agents beliefs. This template is then stored in the agents beliefs so it can be retrieved and used as necessary.

Details of how this template is used can be found in the second snippet of code shown in Listing \ref{code:tower2}. Here, there are a number of plans that play specific roles. The first plan triggers the tower building behaviour when block 'c' is added to the table (this is a feature of the UI that is provided with the environment).

As a follow up experiment, we explored a second version of our program that attempted to construct the tower \verb|['b', 'a', 'd']| after block \verb|'d'| was added.  The LLM failed to achieve this, returning a plan that was similar to the plan offered for the tower \verb|['a', 'b', 'c]|. At this point, we decided not to pursue a more complex experiment where the LLM had to contend with blocks that were on top of one another rather than simply being on the table.

\section{Experiences}
\label{sec:experiences}

While developing our library for integrating LLMs with ASTRA, we reached some early conclusions around the use of the technology:

\begin{itemize}
\item \textbf{LLMs are easy to integrate with AOP languages:} Many libraries exist that make incorporating LLMs easy. The library we chose (Langchain4j) allowed us to do this in a few lines of code.  It was also relatively simple to extract knowledge from beliefs to incorporate into prompts and to generate outputs (plans) in a form that could be consumed by the agent.

\item \textbf{Agentic Workflow Patterns can be realised through existing MAS technologies:} Many of the workflow patterns presented for use in Agentic systems are quite trivial and are easy to implement using traditional agent toolkits.  This is illustrated through the Travel Planner example (Section \ref{sec:travel}) where a round robin group chat model was implemented using the FIPA ACL Request Protocol. While this is not the best fit, it works, and there are many different ways to implement compatible solutions (for example integration with Message Brokers and use of Shared Artifacts).

\item \textbf{Not all workflows require multiple agents:} The use of multiple agents to implement Agentic workflows could be overly complicating the design as most can be implemented through simple plans within a single agent. This is illustrated through the \verb|Reflective...Player| Tic-Tac-Toe example (Section \ref{sec:ttt}), where a single agent implements the expert-assessor workflow.
 
\item \textbf{LLMs are not good at contextual decision making:} It is clear from the limited experiments we have conducted that LLMs are not consistent decision makers. For example, in Tic-Tac-Toe the LLM would fail to detect both losing and winning positions (Section \ref{sec:ttt}). Similarly, in Towerworld (Section \ref{sec:towerworld}), the LLM was not able to work out what action to perform next given the current state and a goal state.

\item \textbf{LLMs do not seem to be good at (complex) reasoning:} In our experiments, the LLM was not good at complex reasoning. It was unable to force a draw in Tic-Tac-Toe (against a linear player) or produce a plan that allowed it to construct any tower other than the tower \verb|['a', 'b', 'c']| in Towerworld (see Section \ref{sec:towerworld}). These results compliment the recent findings of Apple Researchers who explored reasoning about Tower of Hanoi \cite{apple2025}.

\item \textbf{Prompting is a dark art:} Coming up with effective prompts can be challenging and small changes can have a big impact on the quality of the results. While we took time developing the prompts presented in the examples, it is not clear still whether or not we can improve on our results.

\end{itemize}

\section{Related Work}
\label{sec:related}

A recent survey of the state of the art in engineering MAS, which includes a review of works that combine LLMs with agent toolkits (many of which are based on the BDI architecture), shows that there is little work to date on integrating LLMs into agent toolkits like ASTRA \cite{thisVolumeBriolaEtAl}. Perhaps the closest is the proposed extension to the JadexV platform to incorporate LLMs for planning, rule revisions and interaction with human users --- although this is not currently implemented. In terms of the human-user interaction angle, ChatBDI \cite{gatti2025chatbdi} uses LLMs as an interface between human users and BDI agents.  Similarly, \cite{frering2025integrating} explores the use of LLMs for Human-Robot interactions and explainable AI. Finally, \cite{ichida2024bdi} explores the creation of a BDI reasoning cycle that is underpinned by an LLM. This differs from the work presented in this paper, which focuses on integrating LLM calls into classical agent-oriented software engineering and a traditional agent toolkit --- rather than proposing new approaches.

\section{Conclusions}
\label{sec:conclusions}

This paper presents a prototype library for integrating LLMs into a traditional agent programming language, ASTRA (see Section \ref{sec:library}). We used this library to explore three distinct scenarios (Section \ref{sec:examples}): a travel planner example from the LLM community, Tic-Tac-Toe, and Towerworld. Section \ref{sec:experiences} highlights our experiences in creating both the library and the example programs. Overall, it is clear that LLMs can easily be integrated into agent toolkits and that those toolkits can be used in similar scenarios to those described by the Agentic AI community.

A major issue is the quality of reasoning exhibited by LLMs and the challenge of writing effective prompts that can extract useful, relevant and consistent information from them. It is becoming increasingly clear that LLMs are not good at complex reasoning. For example, \cite{topsakal2024evaluating} presents a set of benchmark tests for LLMs playing board games that demonstrate poor capabilities. Similarly, Apple Research recently released a paper discussing the reasoning capabilities of LLMs based on their ability to solve Tower of Hanoi problems \cite{apple2025}. Tower of Hanoi is similar to Towerworld although more challenging due to limitations on the number of pegs available. It is interesting to see that their approach, which is to generate a plan to get from a start state to a goal state is similar to the approach used in Section \ref{sec:towerworld}, as are their findings.

\bibliographystyle{splncs04}
\bibliography{references}
\end{document}